\documentclass[sigconf]{acmart}

\usepackage{amsmath}
\usepackage{multirow}
\usepackage{graphicx}
\usepackage{subcaption}
\usepackage[utf8]{inputenc}
\usepackage{verbatim}
\usepackage{caption}

\usepackage{listings}
\usepackage{xcolor}
\usepackage[acronym,toc]{glossaries}

\newcommand{\todo}[1]{\textcolor{red}{\textbf{[TODO: #1]}}}

\lstdefinelanguage{yaml}{
  morekeywords={topology:,algorithm:,global_rounds:,defaults:},
  keywordstyle=\color{blue}\bfseries,
  basicstyle=\ttfamily\scriptsize,
  comment=[l]{\#},
  commentstyle=\color{gray},
  stringstyle=\color{teal},
  moredelim=[l][\color{magenta}]{---},
  moredelim=[l][\color{magenta}]{...},
  literate=
   *{:}{{{\color{red}:}}}{1}%
    {-}{{{\color{red}-}}}{1}%
}

\lstdefinestyle{yamlstyle}{
  language=yaml,
  frame=single,
  numbers=left,
  xleftmargin=2em,
  breaklines=true,
  breakatwhitespace=false,
  backgroundcolor=\color{yellow!15}
}

\graphicspath{{./}}

\AtBeginDocument{%
  }

\makeglossaries

\newglossaryentry{omnifed}{
  name={\textsf{OmniFed}},
  description={A modular framework for configurable \gls{fl} from edge to \gls{hpc}}
}

\newacronym{ai}{AI}{Artificial Intelligence}
\newacronym{ml}{ML}{Machine Learning}
\newacronym{dl}{DL}{Deep Learning}
\newacronym{nn}{NN}{Neural Networks}
\newacronym{dnn}{DNN}{Deep Neural Network}

\newacronym{fl}{FL}{Federated Learning}
\newacronym{cl}{CL}{Collaborative Learning}
\newacronym{tff}{TFF}{TensorFlow Federated}

\newacronym{hpc}{HPC}{High Performance Computing}
\newacronym{api}{API}{Application Programming Interface}

\newacronym{grpc}{gRPC}{Google Remote Procedure Call}
\newacronym{mpi}{MPI}{Message Passing Interface}
\newacronym{mqtt}{MQTT}{Message Queuing Telemetry Transport}
\newacronym{amqp}{AMQP}{Advanced Message Queuing Protocol}
\newacronym{nccl}{NCCL}{NVIDIA Collective Communications Library}

\newacronym{dp}{DP}{Differential Privacy}
\newacronym{he}{HE}{Homomorphic Encryption}
\newacronym{sa}{SA}{Secure Aggregation}
\newacronym{smpc}{SMPC}{Secure Multi-Party Computation}
\newacronym{tee}{TEE}{Trusted Execution Environment}
\newacronym{hmac}{HMAC}{Hash-based Message Authentication Code}

\newacronym{lr}{LR}{Learning Rate}
\newacronym{sgd}{SGD}{Stochastic Gradient Descent}
\newacronym{dgc}{DGC}{Deep Gradient Compression}
\newacronym{qsgd}{QSGD}{Quantized Stochastic Gradient Descent}

\newacronym{gbdt}{GBDT}{Gradient Boosting Decision Tree}

\copyrightyear{2025}
\acmYear{2025}
\setcopyright{none} % or use {rightsretained} or other appropriate options
\acmConference[SC workshops '25]{Proceedings of Your Conference Name}{Nov. 16--21, 2025}{St. Louis, MO}
\acmBooktitle{Proceedings of The International Conference for High Performance Computing, Networking, Storage, and Analysis}
\acmDOI{XXXXXXX}
\acmISBN{978-1-4503-XXXX-X/25/06}

\author{Sahil Tyagi\textsuperscript{*} \quad Andrei Cozma\textsuperscript{*} \quad Olivera Kotevska \quad Feiyi Wang}
\email{[tyagis, cozmaal, kotevskao, fwang2]@ornl.gov}
\affiliation{%
  \institution{Oak Ridge National Laboratory}
  \country{USA}
}

\thanks{* These authors contributed equally to this work.}

\begin{document}
\title{\gls*{omnifed}: A Modular Framework for Configurable \acrlong*{fl} from Edge to \acrshort*{hpc}}

\begin{abstract}
\gls{fl} is critical for edge and \gls{hpc} where data is not centralized and privacy is crucial.
%
% We present OmniFed, a modular PyTorch framework for federated/collaborative learning that spans edge devices to HPC and cross-facility settings. Users declare topologies as graphs (centralized, hierarchical, etc.) and can mix communication backends (e.g., MPI and gRPC) within a single deployment. 
We present \textsf{OmniFed}, a modular framework designed around decoupling and clear separation of concerns for configuration, orchestration, communication, and training logic.
Its architecture supports configuration-driven prototyping and code-level override-what-you-need customization.
%enabling flexibility without interfering with the core system.
% It provides built-in support for centralized and hierarchical/hybrid topologies and multiple communication mechanisms, such as MPI and gRPC, while treating both as first-class extension points. 
% OmniFed includes built‑in support for various topology templates (e.g., centralized and hierarchical/hybrid) and communication mechanisms (e.g., MPI and gRPC), treating both topologies and communication as first-class extension points within the architecture. 
We also support different topologies, mixed communication protocols within a single deployment, and popular training algorithms.
It also offers optional privacy mechanisms including \gls{dp}, \gls{he}, and \gls{sa}, as well as compression strategies.
% Users can define custom topologies beyond the built-in templates, implement custom communicators, introduce new aggregation strategies, or create custom training loops, all through well-defined extension points that preserve flexibility without interfering with the core system. 
%
%These capabilities are exposed through well-defined extension points, allowing users to customize at multiple levels: topology and orchestration (e.g., new topologies or communication protocols), learning logic (e.g., new aggregation strategies or training loops), and plugins (e.g., new privacy or compression methods), all while preserving the integrity of the core system.
These capabilities are exposed through well-defined extension points, allowing users to customize topology and orchestration, learning logic, and privacy/compression plugins, all while preserving the integrity of the core system.
% This unified, extensible structure improves readability and reproducibility, supports configuration‑driven workflows for rapid prototyping, and enables code‑level override‑what‑you‑need customization via lifecycle hooks, ensuring flexibility without interference. 
% By combining these capabilities in a unified, self-contained architecture, OmniFed improves readability and reproducibility, supports configuration-driven prototyping, and enables code-level override-what-you-need customization via lifecycle hooks.
%
%We evaluate OmniFed on 16 clients on a single 8\(\times\)H100 node in a controlled single-facility setting for reproducibility and performance isolation, using 10+ federated algorithms, and measure convergence, runtime, and overhead, including streaming-data scenarios.
We evaluate multiple models and algorithms to measure various performance metrics.
By unifying topology configuration, mixed-protocol communication, and pluggable modules in one stack, \textsf{OmniFed} streamlines \gls{fl} deployment across heterogeneous environments.
Github repository is available at \href{https://github.com/at-aaims/OmniFed}{\textsf{https://github.com/at-aaims/OmniFed}}.

\end{abstract}


\maketitle

\section{Introduction}\label{sec:intro}

% Reset glossary first-use flags so acronyms expand to full forms again in main body
\glsresetall

As data becomes increasingly distributed, sensitive, and voluminous, conventional centralized \gls{ai} pipelines are no longer practical today.
\Gls{fl} and \gls{cl} techniques developed over the last decade are crucial to develop \gls{dnn} models that learn not just from open, publicly available data, but also from private, sensitive data residing at restricted endpoints.
Algorithms like Federated Averaging~\cite{fedavg} train over devices with limited compute, skewed data quality/quantity, and slow networks.
From a distributed computing perspective, this intuition involves moving the computation instead of moving data~\cite{mapreduce, sparkRDD}.
\gls{fl} emphasizes privacy-preserving training over sensitive or regulated datasets such that data cannot be relocated or reconstructed via model or output inversion attacks.
On the other hand, \gls{cl} focuses on decentralized, in-situ model training where privacy may be less critical, but data locality and institutional boundaries remain key drivers.
Together, \gls{fl}/\gls{cl} offer a transformational opportunity for the next generation of secure and distributed \gls{ai} where models are trained or fine-tuned across institutional, geographic, and disciplinary boundaries without relocating the data.
Development of novel \gls{fl}/\gls{cl} systems and training strategies is thus crucial for scientific collaborations, national security applications, and industrial partnerships.
Training models where the data resides: on partner/shared systems, scientific instruments, and secure edge devices, can help advance both open science and protected interests/missions simultaneously.

We present \gls{omnifed}, a Python-based modular, extensible, configurable, and open-source framework built atop PyTorch~\cite{pytorch} with the goal of enabling \gls{fl}/\gls{cl} from the edge to \gls{hpc} systems, spanning scientific and healthcare instruments, edge computing environments, cloud resources, and large-scale \gls{hpc} systems.
Built with layered abstractions and clear separation of concerns, \gls{omnifed} works in a plug-and-play and override-what-you-need manner via lifecycle hooks.
Hiding the complexity of training topology and underlying communication allows rapid prototyping of new \gls{fl}/\gls{cl} algorithms without excessive boilerplate.
Thus, the end \gls{ml} user focuses solely on training logic (like algorithm, model architecture, loss computation, optimization strategy, etc.) while the framework handles low-level orchestration details and scaffolding.
\gls{omnifed} makes it easy to implement and deploy complex, custom training topologies (such as centralized, decentralized, hierarchical, etc.) via Hydra configuration management~\cite{hydra, omegaconf}.
It also comes with built-in privacy features like \gls{dp}~\cite{diffprivacy2}, \gls{he}~\cite{he1} and \gls{sa}~\cite{secagg1}, gradient compression techniques~\cite{gravac, tyagioverviewpaper}, streaming simulation for real-time learning~\cite{scadles}, and 10+ \gls{fl} algorithms.
The layout of this article is as follows: \S \ref{sec:bgrelatedwork} describes prior works in the \gls{fl}/\gls{cl} space and the need for secure, distributed \gls{ai} for collaborative science, healthcare, finance applications, then \S \ref{sec:design} explains the design principles and core components of \gls{omnifed} along with preliminary results, and \S \ref{sec:conclusion} describes our current and future directions of our work.

\section{Challenges and Related Work}\label{sec:bgrelatedwork}

\subsection{Need for Federated \acrshort*{ai}/\acrshort*{ml}
  Systems}

Language and reasoning models already trained on publicly available internet-scale knowledge have reduced the need for pre-training generic \gls{dnn}s.
The next frontier of computing lies in leveraging these models for private, sensitive, restricted, or distributed datasets for tasks like scientific discovery, healthcare, finance, and other critical applications.
Further training or fine-tuning models over such datasets allows for personalized, context-aware intelligence for individual users~\cite{appleintelligence}, and also help organizations working at the intersection of \gls{hpc}, security, and scientific discovery.
The decentralized nature of \gls{fl}/\gls{cl} offers a new approach to scientific workflows that have prior relied on centralized data aggregation and monolithic data pipelines.
The ever-increasing volume, sensitivity, and distribution of scientific data, ranging from neutron scattering experiments~\cite{neutronscatterexp} to genomic repositories~\cite{genomicrepo}, have created technical, logistical, and ethical barriers to centralization, thus making it infeasible or prohibited.
In the realm of open models and sciences, \gls{fl} enables collaborative \gls{ai}/\gls{ml} across research institutions, laboratories, academia, and industry while preserving data governance.
This supports efforts like climate modeling, materials science, cyber-threat detection, and energy infrastructure monitoring, while retaining local data control.
The integration of \gls{fl}/\gls{cl} with edge, cloud, and \gls{hpc} systems thus creates a new capability tier of privacy and security-aware distributed intelligence operating across layers of the scientific enterprise.

In the context of national labs and computing facilities, \gls{fl} opens up new frontiers in scientific modeling by combining federated models with \gls{hpc}-powered simulations and developing self-updating digital twins~\cite{brewerdigitaltwin} for nuclear reactors, electric grids, materials, and climate systems.
The convergence of \gls{fl} and simulations enables adaptive, real-time modeling at scale, offering breakthrough capabilities for monitoring, control, and prediction in complex systems.

\begin{figure*}
  \centering
  \subfloat[Centralized]{%
    \includegraphics[width=0.3\columnwidth]{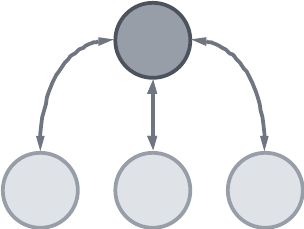}
    \label{fig:central}
  }\hfill
  \subfloat[Ring]{%
    \includegraphics[width=0.33\columnwidth]{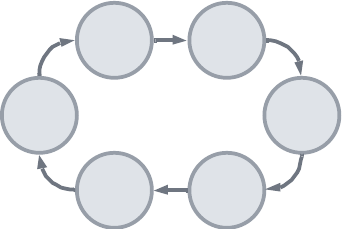}
    \label{fig:decentralring}
  }\hfill%\\[0.5em]
  \subfloat[Peer-to-Peer]{%
    \includegraphics[width=0.3\columnwidth]{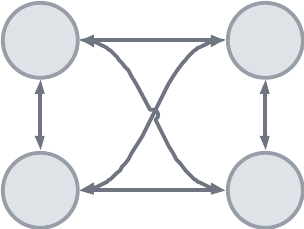}
    \label{fig:decentralp2p}
  }\hfill
  \subfloat[Hierarchical Tree]{%
    \includegraphics[width=0.4\columnwidth]{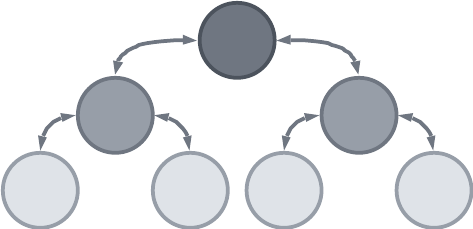}
    \label{fig:hierarchicaltree}
  }
  \caption{Participants connected under different topological arrangements or configurations (central, ring, p2p, and hierarchical).}
  \label{fig:clustertopology}
\end{figure*}

\subsection{Current Frameworks and Limitations}

Many frameworks have been developed in recent years, each with distinct pros and cons.
For instance, \gls{tff}~\cite{tff} developed by Google is Python-based with tight TensorFlow~\cite{tensorflow} integration to simulate \gls{fl} experiments.
Although \gls{tff} is excellent for research and experimentation, it lacks scalability so it may not be ideal for real-world deployment.
NVFLARE~\cite{nvflare} developed by NVIDIA is an active and well-documented \gls{fl} framework supporting TensorFlow and PyTorch that also supports privacy features.
However, it faces scalability challenges where performance may drop as the number of active clients rises.
%and lacks support for heterogeneous clients.
%and does not support on-device training at the edge.
Communicating large models between clients and server poses scalability challenges with \gls{grpc} protocol~\cite{grpc}, thus demanding high memory/network bandwidth for data exchange.
Flower~\cite{flower} is language and \gls{ml}-backend agnostic, and supports both simulations and real-world deployments.
However, it can require manual orchestration for advanced deployments.
Although privacy techniques like \gls{sa} have been integrated, privacy tools are still in development, and advanced use cases may require extra customization.
Flower lacks advanced production tools for monitoring and orchestration, and supports synchronous training by default, which can be prone to straggler slowdowns~\cite{tyagisharmataming, omnilearn}.

OpenFL~\cite{openFL} by Intel is a Python-based \gls{fl} framework designed for multi-institution settings, optimized for secure and cross-org \gls{fl} in regulated industries via \gls{tee}.
However, it only supports centralized server setup (prone to a single point of failure), is built mainly around Intel platforms, lacks built-in privacy tools, and has demonstrated limited mobile/edge use cases.
OpenFL does not support decentralized or asynchronous training.
Medical Open Network for \gls{ai} (MONAI)~\cite{monai} is a PyTorch-based, domain-optimized framework for \gls{fl} over medical and healthcare applications, and uses NVIDIA-based tools for distributed, mixed-precision training.
However, MONAI is specialized for medical imaging and lacks direct support for other healthcare domains (e.g., genomics), optimized primarily for NVIDIA GPUs, and may require additional support for integration with clinical systems.
PySyft~\cite{pysyft} by OpenMined treats privacy as a first-class citizen for Python-based \gls{ai}/\gls{ml} workflows using PyTorch and TensorFlow.
It contains a rich privacy toolkit with features like \gls{he}, \gls{dp}, and \gls{smpc}, as well as built-in access controls.
However, PySyft requires additional setup and lacks built-in orchestration for \gls{fl} workflows.
As a standalone, PySyft runs in simulation mode only.
Running \gls{fl} at scale requires additional integration with other OpenMined components like PyGrid, which may make real-world deployment non-trivial.
%Further, the framework can be slower to run for large-scale applications.

FedML~\cite{fedML} is Python-based, framework agnostic, and optimized for cross-platform \gls{fl} leveraging cloud and on-device learning over edge/mobile devices.
It also encapsulates various privacy-preserving techniques and trusted execution to mitigate data leakage risks.
However, FedML can face scalability issues in large-scale training and struggle with data or systems heterogeneity.
APPFL~\cite{appfl1, appfl2} developed at Argonne National Laboratory is designed for research institutions running cross-silo experiments.
The framework is developed with a modular design, supports advanced \gls{fl} algorithms, has built-in privacy features like \gls{dp}, and supports communication backends like \gls{mpi}~\cite{mpicollectives}, \gls{grpc}, and Globus Compute~\cite{globus}, supports synchronous/asynchronous scheduling and lossy compression~\cite{tyagioverviewpaper} to reduce communication overhead.
%However, with its modular architecture comes complex configuration overhead, and it has various dependency requirements that may make deployment non-trivial in complex environments.
However, its dependency requirements may make deployment non-trivial in complex environments.
IBM \gls{fl}~\cite{ibmFL} is an enterprise-grade framework with visualization dashboards that enables application monitoring, comes with various security tools and advanced aggregation methods.
It's also framework-agnostic and uses communication protocols like \gls{grpc} and WebSockets.
However, IBM \gls{fl} is not entirely open-source, and production-grade access comes through commercial offerings.
The integrated tools can be heavyweight and potentially make deployment cumbersome and complex.

Flotilla~\cite{flotilla} supports both PyTorch and TensorFlow, works with heterogeneous edge devices, and supports advanced \gls{fl} algorithms.
The system is less focused on privacy and security, and deployment orchestration can be complex in distributed settings.
Apple's private \gls{fl} (pfl)~\cite{applePFL, granqvist2024pfl} is a privacy-centric framework that supports PyTorch, TensorFlow, and non-neural models like \glspl{gbdt}.
However, the toolkit is designed for simulations instead of direct production deployments.
It also has limited portability outside of Apple's ecosystem as it uses proprietary infrastructure like secure enclaves.
FATE~\cite{fateFL}, developed by China-based WeBank, supports industrial-grade \gls{fl} with secure computation features built into it that make it ideal for finance and enterprise use.
The framework runs on standalone systems, Docker~\cite{docker}, Kubernetes~\cite{KubernetesProject}, and Spark~\cite{sparkRDD}.
However, setup can be challenging and complex with all its dependencies initially.
These frameworks have different strengths and weaknesses, and may trade off between ease-of-use and scalability, privacy, or feature-rich orchestration.
We develop \gls{omnifed} to provide users with clear-cut abstractions to choose between these trade-offs in a modular fashion with minimal code change.
By hiding away the scaffolding and orchestration details, \gls{fl} researchers can focus on developing algorithms and optimization strategies to produce secure and quality models. %with fast training.

\section{\gls*{omnifed}
  Framework}\label{sec:design}

\subsection{Motivation}\label{subsec:designmotivation}

We develop \gls{omnifed} with the goal to enable researchers and developers to easily evaluate different \gls{fl}/\gls{cl} algorithms, as well as permit rapid prototyping of new learning algorithms without excessive boilerplate.
For instance, switching from FedAvg~\cite{fedavg} to FedProx~\cite{fedprox} should ideally be a single line change that allows users to use different algorithms.
Additionally, most frameworks assume a centralized training topology (shown in Figure~(\ref{fig:central})), so implementing custom learning strategies across skewed and unbalanced islands of heterogeneous clients or levels of topology (e.g., hierarchical tree in Figure~(\ref{fig:hierarchicaltree})) can be non-trivial and even complex.
Federated nodes/devices may be connected in different topologies, as shown in Figure~(\ref{fig:clustertopology}), each offering distinct trade-offs.
For instance, a centralized or star topology comprises a single coordinator that communicates with all participants, or a decentralized topology where a node is connected only to its immediate neighbors, or a peer-to-peer topology where every participant can directly communicate with every other node.
A hierarchical tree consists of a multi-level structure with parent-child aggregation paths.
At the same time, a hub-and-spoke topology may contain islands of densely connected nodes, with sparse connections across regions/facilities.
Thus, a flexible \gls{fl} framework should allow both simulation and real deployment across different clusters and training configurations.

\subsection{Design Philosophy and Principles}\label{subsec:designvision}

\gls{omnifed} is developed to treat modularity, flexibility, and extensibility as first-class citizens in the \gls{fl}/\gls{cl} ecosystem.
With precise, layered abstractions, local computation, communication logic, and algorithmic control strategy are well separated and can easily be swapped out with \gls{omnifed}'s configuration-driven, schema-based workflow that is agnostic to topology (such as those shown in Figure~(\ref{fig:clustertopology})) and communication protocols (e.g., \gls{grpc}~\cite{grpc}, \gls{mpi}~\cite{mpicollectives}, \gls{mqtt}~\cite{mqtt}).
With distinct and minimal interfaces, plug-and-play components, and an override-what-you-need paradigm, users can add or remove different privacy features and compression techniques with minimal code change.
We provide single-file algorithm plugins where existing \gls{fl} techniques can be evaluated with a line change in a job's YAML configuration file.
Additionally, new \gls{fl}/\gls{cl} algorithms can be easily prototyped by extending the base algorithm class and overriding abstract methods.
%like \texttt{train} and \texttt{aggregate}.
With a developer-friendly, consistent, and intuitive usage pattern, \gls{omnifed} allows users to focus on core tasks instead of managing the scaffolding and setup around a job.

\subsection{Core Components}\label{subsec:flyracomponents}

We implement \gls{omnifed} in Python with PyTorch~\cite{pytorch} as the core \gls{ai}/\gls{ml} backend, and use YAML-based configuration management with Hydra~\cite{hydra, omegaconf}, a framework for complex \gls{ml} and data-science workflows that enables hierarchical composition, command-line overrides, and flexible job deployment.
Jobs are deployed for distributed execution over Ray~\cite{rayAI, ray_ai_anyscale}, an \gls{ai} compute engine for orchestrating distributed workloads across heterogeneous hardware.
Ray also supports fine-grained parallelism across CPUs and GPUs, allowing users to easily scale jobs from a local machine to multi-node clusters.
Based on this stack, \gls{omnifed}'s core modules are:

\begin{itemize}
  \item \textit{\textbf{Engine}} is responsible for handling the orchestration rounds, resources, metrics, etc.

  \item \textit{\textbf{Topology}} defines the node graph and coordination patterns.

  \item \textit{\textbf{Node}} corresponds to a \gls{fl}/\gls{cl} participant that may hold a model, training data, and communicators.
        Based on the role of the node in executing a specific algorithm, it can either be a client, an aggregator or a relay.

  \item \textit{\textbf{Communicator}} comprises of a unified \gls{api} with abstract primitives that use different communication protocols under the hood (for e.g., \gls{grpc}~\cite{grpc}, \gls{mpi}~\cite{mpicollectives}, \gls{mqtt}~\cite{mqtt}, etc.).

  \item \textit{\textbf{Algorithm}} defines the local training logic and overall learning strategy through lifecycle hooks.
        Users can choose from a suite of built-in ones (see \S \ref{subsub:switchAlgos}) or implement their own.
\end{itemize}

The \textit{Engine} is the core orchestrator for all \gls{fl}/\gls{cl} experiments.
It's responsible for launching and coordinating all distributed experiments, managing node lifecycle and resource allocation, and collecting report metrics and statistics.
The \gls{omnifed} engine is instantiated via a Hydra-based YAML configuration file, and coordinates with \textit{Topology} definitions to spawn \textit{Node} actors (using Ray~\cite{rayAI}) and manage \textit{Algorithm} implementations.

The training \textit{Topology} defines the structure, coordination, and scheduling patterns for distributed nodes.
It's responsible for defining the node graph structure and relationships between nodes, supports templates for centralized, decentralized, and hierarchical topologies (as shown in Figure~(\ref{fig:clustertopology})).
%Additionally, custom and complex topologies can easily be defined via \textit{Topology}'s graph-based representations from the job's YAML configuration, which is then used by the engine to spawn and organize \textit{Node} actors.
We are working on developing custom and complex topologies via \textit{Topology}'s graph-based representations from the job's YAML configuration, to be used by the engine to spawn and organize \textit{Node} actors.
%It also determines how nodes communicate with each other (i.e., edges of the graph).
The edges of the graph will determine which nodes can communicate with each other.

\textit{Node} can be any participant in the federation.
It's responsible for managing local model state, data, and device/client resources, interfaces with the \textit{Communicator} for data exchange, and executes the local \textit{Algorithm} implementations.
\textit{Nodes} are instantiated as Ray actors by the \gls{omnifed} \textit{Engine} and configured via Hydra YAML files with optional per-node overrides.
A \textit{Node} closely coordinates with \textit{Algorithm} components for local compute logic like train, eval, etc.
Currently, \textit{Node} can serve the role of a trainer or aggregator.
The former performs local training on data, while the latter collects, merges, and redistributes model updates.
%We plan to add more specific roles like a \textit{relay} to forward messages between multi-tier topologies (e.g., edge $\rightarrow$ aggregator $\rightarrow$ cloud), or \textit{coordinator} to manage execution across different sub-groups (e.g., clusters within a hierarchy).
%Additionally, roles are composable per-node and \textit{Nodes} can play multiple roles (e.g., a node can both train and aggregate).
Every \textit{Node} implements an inner communicator by default for its topology.
To implement hierarchical \gls{fl}, \textit{Nodes} additionally implement an outer communicator to exchange data across sub-topologies.

\textit{Communicator} is needed for data exchange among \textit{Nodes} like model parameters, gradients, control signals, etc.
The module exposes a unified \gls{api} that abstracts the underlying protocol while standardizing data exchange operations with consistent behavior and configurable backend selection without code changes.
We currently implement \gls{mpi} collectives via \texttt{TorchDistCommunicator} using PyTorch Distributed (so works with various underlying libraries such as \gls{nccl}~\cite{nccl}, \gls{mpi}~\cite{mpich}, or Gloo~\cite{gloo}).
%A \texttt{GrpcCommunicator} uses \gls{grpc} and protocol buffers~\cite{protobuf} for distributed \gls{fl}/\gls{cl} scenarios where a \gls{grpc} server receives, aggregates, and broadcasts updates sent by clients over heterogeneous networks.
A \texttt{GrpcCommunicator} based on \gls{grpc} and protocol buffers~\cite{protobuf}, has a server that receives, aggregates, and broadcasts updates sent by clients over heterogeneous networks in distributed \gls{fl}/\gls{cl}.
For middleware systems, we are additionally developing \gls{amqp}-based~\cite{amqpprotocol} \texttt{AMQPCommunicator} to support publish-subscribe message patterns where clients push updates to a queue, which is subsequently pulled by the aggregator \textit{Node}.

\textit{Algorithm} provides \gls{fl} logic through configurable lifecycle hooks, and called by the \textit{Node} at various stages, such as local training, model update, post-aggregation processing, etc.
This abstraction separates concerns through well-defined interfaces; \textit{Node} manages resource and data, while \textit{Algorithm} provides the learning strategy.
The module is flexible so that different implementations can override only what they need, and extensible to support diverse \gls{fl}/\gls{cl} paradigms and topologies.
A Hydra-based YAML file describes the simulation or deployment configuration.
\gls{omnifed} comes with pre-built \textit{Topology} templates that enable quick and straightforward configurations for common patterns like centralized, decentralized, and hierarchical.
%A custom graph can be implemented, which explicitly defines bespoke nodes and edges.
%One can also implement a custom graph with explicitly defined nodes and edges.
We are actively working on implementing custom graphs with explicitly defined nodes and edges.
Each \textit{Node} contains the following information: role, \textit{Communicator}, PyTorch model, data, and \textit{Algorithm}, while a \textit{Topology} describes a set of \textit{Nodes} and edges, and an \textit{Engine} is an orchestration point for \textit{Topology}.
We also integrate various privacy-preserving techniques into \gls{omnifed}, like \gls{dp}~\cite{diffprivacy1, diffprivacy2}, \gls{he}~\cite{he1} and \gls{sa}~\cite{secagg1}.
To add \gls{dp}, we integrate with PETINA~\cite{petina}, a library of privacy-preserving algorithms, while \gls{he} is added through Tenseal~\cite{Benaissa2021TenSEALAL} based on Microsoft SEAL~\cite{microsoftSEAL}.
\gls{sa} is currently prototyped with \gls{hmac}~\cite{pythonHMAC} and Hashlib~\cite{pythonHashlib} to generate a shared key between any two clients in a deterministic manner.
We plan to replace this with Diffie-Hellman key exchange~\cite{Diffie1976NewDI}.

\subsection{Configuration and Deployment}\label{subsec:configdeploy}

\begin{comment}
We test over an NVIDIA DGX server with 8 H100 GPUs to demonstrate \gls{omnifed}'s features and ease-of-use.
Models are trained with 16 clients with the following training configurations:

\begin{itemize}
  \item ResNet18~\cite{resnet18} over CIFAR10~\cite{cifar10100datasets} is trained with initial \gls{lr} 0.01, momentum 0.9, weight decay 0.0001, cross-entropy loss, per-node batch-size 32, and \gls{lr} decay of 0.1 after 100, 150, and 200 epochs.

  \item VGG11~\cite{vgg11} over CIFAR100~\cite{cifar10100datasets} is trained with \gls{lr} 0.01, momentum factor 0.9, weight decay 0.0005, node batch-size 32, and \gls{lr} decay 0.2 after 50, 75, and 100 epochs, respectively.

  \item AlexNet~\cite{alexnet} over CalTech101~\cite{caltech101dataset} is trained with \gls{lr} 0.01, momentum 0.9, weight decay 0.0005, batch-size 32, and \gls{lr} decay 0.1 after 25, 50, and 75 epochs.

  \item MobileNetV3~\cite{mobnetv3} over CalTech256~\cite{caltech101dataset} is trained with \gls{lr} 0.1, momentum 0.9 , weight decay 0.0001 and \gls{lr} decay 0.1 after every 40 epochs.
\end{itemize}
\end{comment}

We test with 16 clients over NVIDIA DGX server with 8 H100 GPUs to demonstrate \gls{omnifed}'s features and ease-of-use.

\textit{Training setup}: ResNet18~\cite{resnet18} over CIFAR10~\cite{cifar10100datasets} is trained with initial \gls{lr} 0.01, momentum 0.9, weight decay 1e-4, cross-entropy loss, per-node batch-size 32, and \gls{lr} decay of 0.1 at 100, 150, and 200 epochs.
VGG11~\cite{vgg11} over CIFAR100~\cite{cifar10100datasets} uses \gls{lr} 0.01, momentum factor 0.9, weight decay 5e-4, node batch-size 32, and \gls{lr} decay 0.2 at 50, 75, and 100 epochs respectively.
AlexNet~\cite{alexnet} over CalTech101~\cite{caltech101dataset} trained with \gls{lr} 0.01, momentum 0.9, weight decay 5e-4, batch-size 32, and \gls{lr} decay 0.1 at 25, 50, and 75 epochs.
MobileNetV3~\cite{mobnetv3} over CalTech256~\cite{caltech101dataset} runs with \gls{lr} 0.1, momentum 0.9, decay 1e-4 and \gls{lr} decay 0.1 every 40 epochs.

\subsubsection{Switching between different algorithms}\label{subsub:switchAlgos}
%\textcolor{purple}{easy to evaluate both systems and statistical performance of different \gls{fl} algorithms}

\begin{figure}
  \centering
  \scriptsize
  \begin{lstlisting}[style=yamlstyle]
defaults:
  - override topology: centralized
  - override algorithm: fedavg
  - override model: resnet18
  - override datamodule: cifar10

topology:
  _target_: src.omnifed.topology.CentralizedTopology
  num_clients: 8
  inner_comm:
    _target_: src.omnifed.communicator.GrpcCommunicator
    master_port: 50051
    master_addr: 127.0.0.1

algorithm:
  _target_: src.omnifed.algorithm.FedAvg
  lr: 0.01

global_rounds: 2

\end{lstlisting}
  \normalsize
  \caption{Example config for centralized FedAvg on ResNet18.
    From the YAML config, users can change the algorithm (e.g., FedAvg to FedProx), type of topology (centralized to decentralized), communicator (\gls{grpc} or \gls{mpi}), model, dataset, aggregation frequency/rounds and hyperparameters (step-size).}
  \label{fig:yamlconfigFLAlgos}
\end{figure}

\begin{figure}
  \centering
  % First row
  \subcaptionbox{ResNet18\label{fig:epochres18}}{%
    \includegraphics[width=0.49\columnwidth]{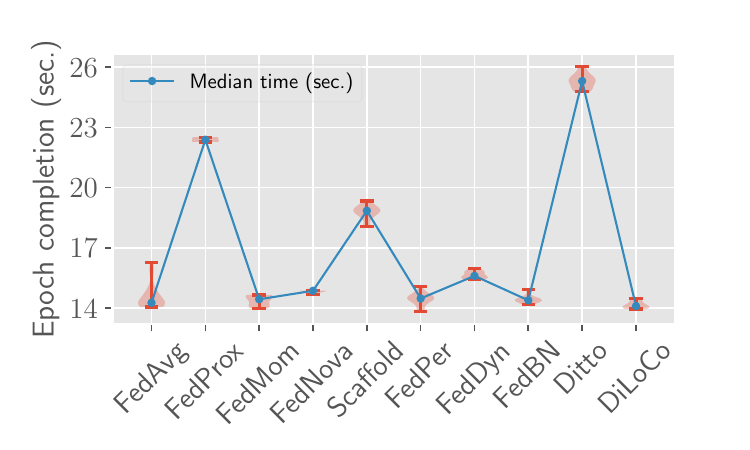}
  }\hfill
  \subcaptionbox{VGG11\label{fig:epochvgg11}}{%
    \includegraphics[width=0.49\columnwidth]{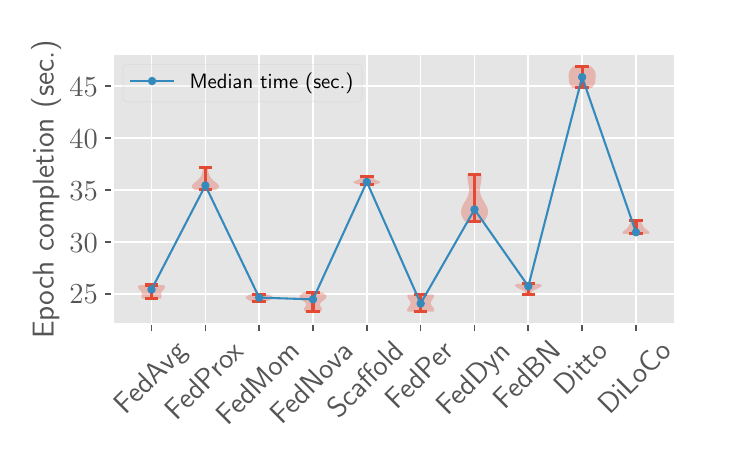}
  }\\[0.5em]
  % Second row
  \subcaptionbox{AlexNet\label{fig:epochalex}}{%
    \includegraphics[width=0.49\columnwidth]{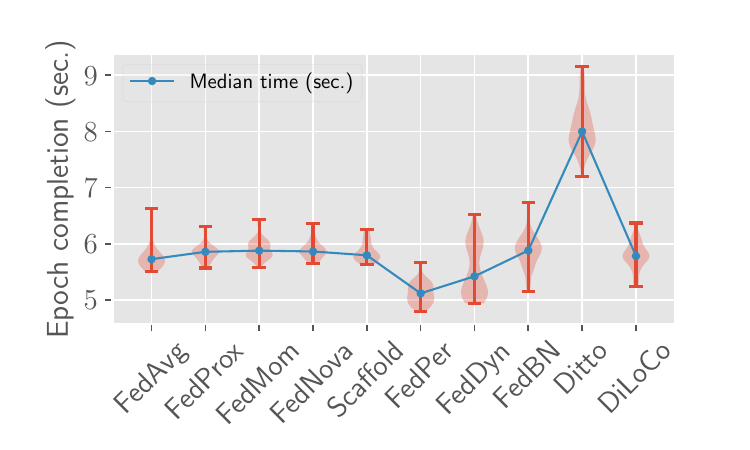}
  }\hfill
  \subcaptionbox{MobileNetV3\label{fig:epochmobv3}}{%
    \includegraphics[width=0.49\columnwidth]{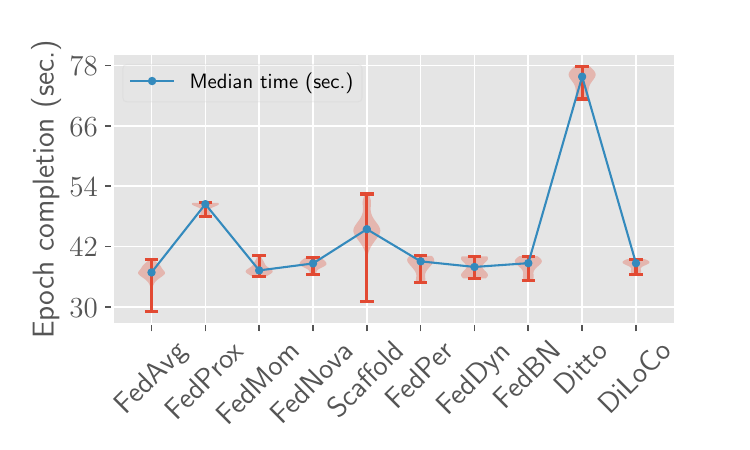}
  }
  \caption{Epoch completion time for different \gls{fl} algorithms.}
  \label{fig:FLAlgoEpochTime}
\end{figure}

\begin{table}
  \centering
  \small
  \setlength{\tabcolsep}{4pt} % Reduce column separation
  \renewcommand{\arraystretch}{1.1} % Slightly increase row height
  \begin{tabular}{|c|c|c|c|c|}
    \hline
    \textbf{Algorithm} & \textbf{ResNet18} & \textbf{VGG11}  & \textbf{AlexNet} & \textbf{MobileNetV3} \\
    \hline
    FedAvg             & 99.32\%           & \textbf{86.6\%} & 87.9\%           & 81.35\% \\
    \hline
    FedProx            & 99.26\%           & 86.31\%         & 87.98\%          & \textbf{82.96\%} \\
    \hline
    FedMom             & 99.14\%           & 66.39\%         & 63.85\%          & 48.98\% \\
    \hline
    FedNova            & 91.18\%           & 14.1\%          & 58.1\%           & 22.27\% \\
    \hline
    Moon               & \textbf{99.46\%}  & 81.67\%         & 87.28\%          & 81.4\% \\
    \hline
    FedPer             & 90.9\%            & 26.93\%         & 82.94\%          & 14.59\% \\
    \hline
    FedDyn             & 99.31\%           & 86.18\%         & \textbf{88.78\%} & 79.15\% \\
    \hline
    FedBN              & 99.33\%           & 86\%            & 88.7\%           & 78.65\% \\
    \hline
    Ditto              & 73.64\%           & 5.5\%           & 40\%             & 9.84\% \\
    \hline
    DiLoCo             & 84.88\%           & 5.1\%           & 45.17\%          & 15.47\% \\
    \hline
  \end{tabular}
  \caption{Convergence quality of various \gls{fl} algorithms.}
  \label{tab:FLAlgoAccuracy}
\end{table}

We have currently implemented the following in \gls{omnifed}'s \textit{Algorithm} module: FedAvg~\cite{fedavg}, FedProx~\cite{fedprox}, FedMom~\cite{fedmom}, FedNova~\cite{fednova}, Scaffold~\cite{scaffold}, Moon~\cite{moon}, FedPer~\cite{fedper}, FedDyn~\cite{feddyn}, FedBN~\cite{fedbn}, Ditto~\cite{ditto} and DiLoCo~\cite{diloco}.
An example Hydra-based YAML configuration for training ResNet18 on CIFAR10 over a centralized topology with \gls{grpc} communicator and FedAvg algorithm is illustrated in Figure~(\ref{fig:yamlconfigFLAlgos}).
The training algorithm can easily be swapped at line 16, and the specific hyperparameters can subsequently be defined in the config.
For e.g., we can change FedAvg to FedProx and add the proximal-term related parameter $\mu$ under the algorithm section.
This makes comparing different \gls{fl}/\gls{cl} algorithms fairly easy with minimal code change, with results shown in Table~(\ref{tab:FLAlgoAccuracy}) and Figure~(\ref{fig:FLAlgoEpochTime}).
The latter logs the average epoch completion time across the four models while former denotes the final test accuracy of various algorithms.
\gls{fl} developers/researchers can thus quickly deploy their applications with different configurations and compare both the systems' and statistical performance of different algorithms.
%It is to be noted that some configs like DiLoCo do not perform well on most of the evaluated models, as the algorithm is optimized for large language models to use AdamW~\cite{adamw} in client-local and momentum \gls{sgd} across clients during the aggregation phase.
%With the default \gls{fl} algorithm hyperparameters set in \gls{omnifed}, ResNet18 achieved the highest accuracy with Moon, VGG11 with FedAvg, AlexNet with FedDyn, and MobileNetV3 with FedProx.
% TODO: performance of other algos MAY BE be improved by adjusting its specific hyperparams. still can easily be done by simply changing the config file.
Some algorithms like DiLoCo are configured for specific settings (e.g., large language models with AdamW~\cite{adamw} for local optimization and momentum \gls{sgd} for cross-client aggregation) and may exhibit sub-optimal performance in other settings.
%without hyperparameter tuning.
We do not perform extensive hyperparameter tuning as the demonstration is intended to validate \gls{omnifed}'s support for diverse algorithms under unified configuration.
In default settings, ResNet18 achieved highest accuracy with Moon, VGG11 with FedAvg, AlexNet with FedDyn, and MobileNetV3 with FedProx.
The current results were obtained with the default parameters specified for an algorithm (see \href{https://github.com/at-aaims/OmniFed}{github}); performance may further be improved via algorithm-specific hyperparameter tuning, which is easily accomplished by modifying the configuration file.
%Performance of individual algorithms could be improved through algorithm-specific hyperparameter tuning, which can be easily accomplished by modifying the configuration file.
\vspace{-2.85em}

\subsubsection{Communication savings via Compression}\label{subsub:compressomnifed}

\begin{figure}
  \centering
  \scriptsize
  \begin{lstlisting}[style=yamlstyle]
inner_comm:
    _target_: src.omnifed.communicator.TorchDistCommunicator
    port: 28670
    compression: 
        _target_: src.omnifed.communicator.compression.TopK
        k: 1000x
\end{lstlisting}
  \normalsize
  \caption{Configuring Communicator with Compression.
    Different compressors (and its related parameters) can be specified from YAML config as part of the communication module.}
  \label{fig:yamlconfigCompress}
\end{figure}
%\vspace{-0.7em}

\begin{figure}
  \centering
  % First row
  \subcaptionbox{ResNet18\label{fig:compressres18}}{%
    \includegraphics[width=0.49\columnwidth]{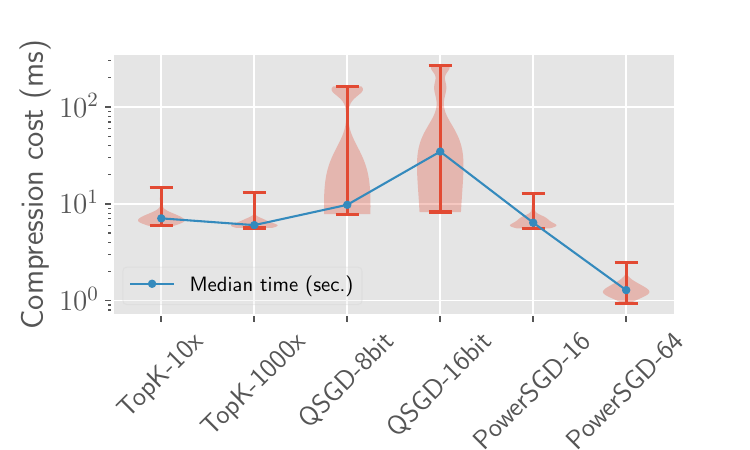}
  }\hfill
  \subcaptionbox{VGG11\label{fig:compressvgg11}}{%
    \includegraphics[width=0.49\columnwidth]{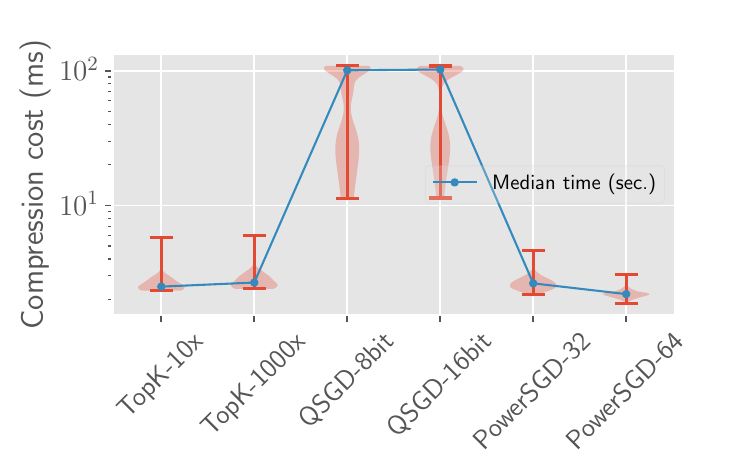}
  }\\[0.5em]
  % Second row
  \subcaptionbox{AlexNet\label{fig:compressalex}}{%
    \includegraphics[width=0.49\columnwidth]{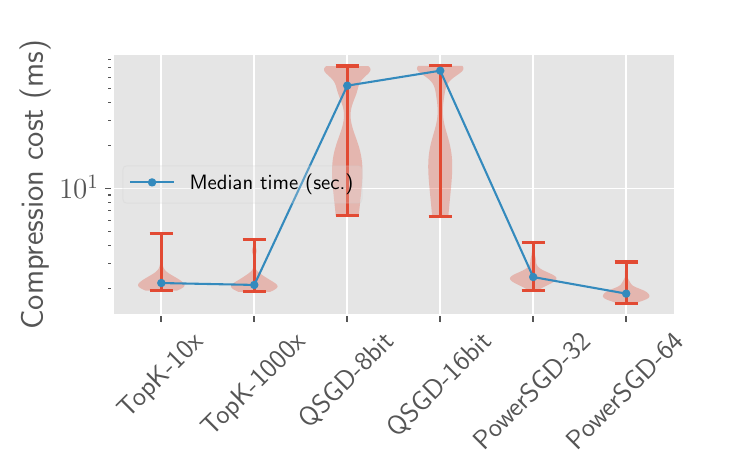}
  }\hfill
  \subcaptionbox{MobileNetV3\label{fig:compressmobv3}}{%
    \includegraphics[width=0.49\columnwidth]{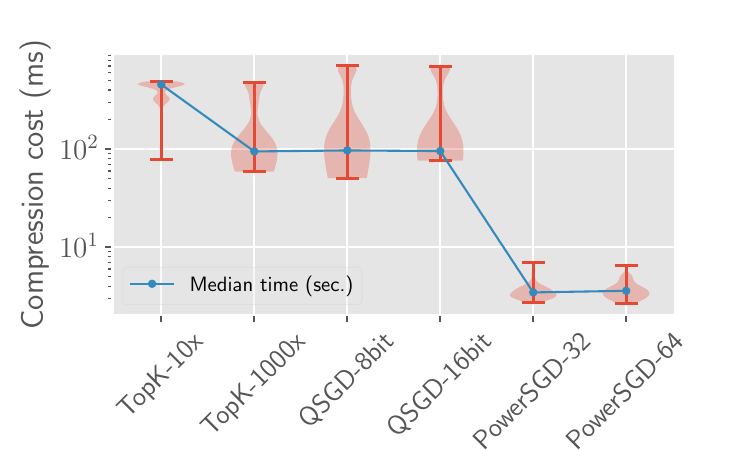}
  }
  \caption{Compression overhead of different techniques.}
  \label{fig:CompressOverhead}
\end{figure}

\begin{table}
  \centering
  \small
  \setlength{\tabcolsep}{4pt} % Reduce column separation
  \renewcommand{\arraystretch}{1.1} % Slightly increase row height
  \begin{tabular}{|c|c|c|c|c|}
    \hline
    \textbf{Compression}         & \textbf{ResNet18} & \textbf{VGG11}   & \textbf{AlexNet} & \textbf{MobileNetV3} \\
    \hline
    Top\textit{k}-10$\times$     & 99.09\%           & 84.6\%           & 87.23\%          & 78.82\% \\
    \hline
    Top\textit{k}-1000$\times$   & 99.05\%           & 78.05\%          & 75.75\%          & 77.56\% \\
    \hline
    \acrshort*{dgc}-10$\times$   & 99.18\%           & 84.22\%          & 87.61\%          & 79\% \\
    \hline
    \acrshort*{dgc}-1000$\times$ & 98.5\%            & 78.75\%          & 76.28\%          & 72.77\% \\
    \hline
    \acrshort*{qsgd} 8-bit       & \textbf{99.26\%}  & 85.48\%          & \textbf{89\%}    & \textbf{79.72\%} \\
    \hline
    \acrshort*{qsgd} 16-bit      & 99.12\%           & \textbf{85.52\%} & 88.3\%           & 77.24\% \\
    \hline
    PowerSGD r-64                & 99.07\%           & 75.45\%          & 75\%             & 77.69\% \\
    \hline
    PowerSGD r-32                & 90.31\%           & 6.74\%           & 40\%             & 77.79\% \\
    \hline
  \end{tabular}
  \caption{Convergence quality in gradient compression.}
  \label{tab:FLCompressAccuracy}
\end{table}

Similar to \S \ref{subsub:switchAlgos}, we compare the overall parallel and statistical performance of various compression techniques.
Currently, we have implemented the following gradient sparsification methods: Top\textit{k}~\cite{topk}, \gls{dgc}~\cite{dgc}, RedSync~\cite{redsync}, SIDCo~\cite{sidco} and Random\textit{k}~\cite{gravac}, \gls{qsgd} quantization~\cite{qsgd} and PowerSGD low-rank approximation~\cite{powersgd}.
A compressor and its parameters can be specified from within the config as part of the communicator, as shown for Top\textit{k} 1000$\times$ in Figure~(\ref{fig:yamlconfigCompress}).
Figure~(\ref{fig:CompressOverhead}) shows the overhead of different compressors at various compression factors, while Table~(\ref{tab:FLCompressAccuracy}) logs the final test accuracy.
Top\textit{k} and \gls{dgc} are compressed to 10$\times$ and 1000$\times$ factors, while \gls{qsgd} compresses to 2$\times$ and 4$\times$ with 16-bit and 8-bit, respectively (compressed w.r.t.
32-bit floats).
The gradients are decomposed to rank 32 and 64 with PowerSGD compression.
Although \gls{qsgd} generally attains better accuracy than other methods (likely from using lower compression factors 2$\times$ and 4$\times$), it comes at the cost of higher compression + communication cost (from Figure~(\ref{fig:CompressOverhead})).
The total overhead of different compressors may vary due to the effective compression factor applied to reduce the volume of updates, and the communication collective used to exchange updates.
Sparsification methods like \gls{dgc} use all-gather, while quantization and low-rank approximation techniques are compatible with all-reduce.
%\vspace{-2em}

\subsubsection{Real-time learning over streaming data}

\begin{figure}
  \centering
  \begin{subfigure}[t]{0.48\columnwidth}
    \centering
    \includegraphics[width=\linewidth]{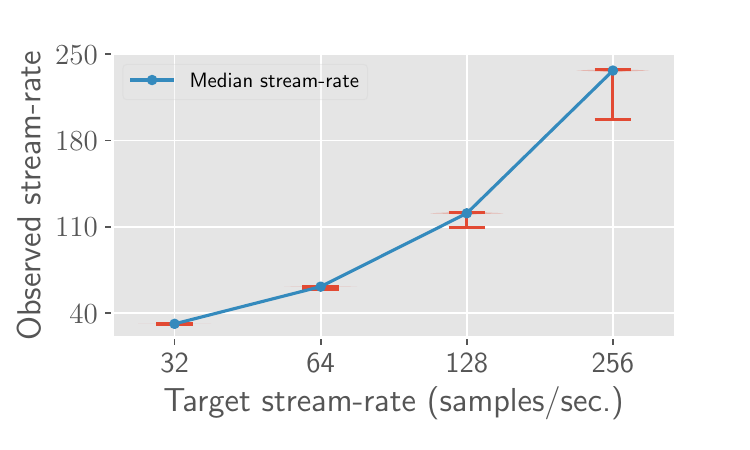}
    \caption{Effective stream-rate}
    \label{fig:multistreamrates}
  \end{subfigure}
  \hfill
  \begin{subfigure}[t]{0.48\columnwidth}
    \centering
    \includegraphics[width=\linewidth]{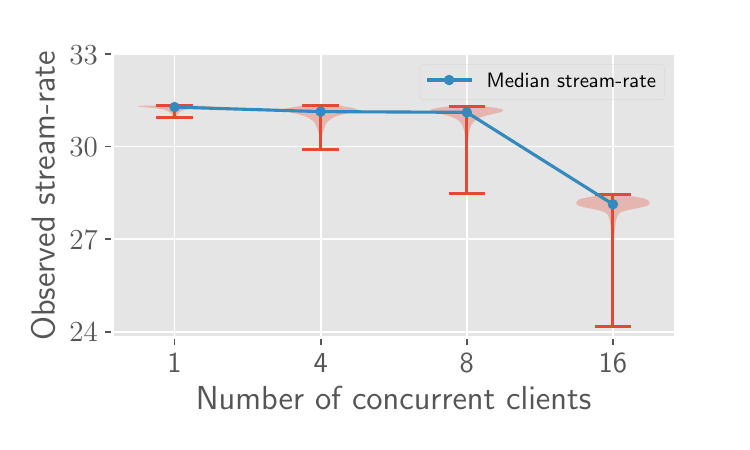}
    \caption{Multi-client stream-rate}
    \label{fig:multiclientstreams}
  \end{subfigure}
  \caption{Streaming simulation for real-time learning.}
  \label{fig:streamingSim}
\end{figure}

To simulate settings with streaming data, we integrate \gls{omnifed} with Apache Kafka~\cite{apachekafka} to parse datasets that are subsequently published to Kafka topics.
Kafka handles ordering issues within a partition, and can be paired with streaming frameworks like Apache Flink \cite{apacheflink} to handle late or out-of-order events.
The data streamed to a specific client (corresponding to its topic) can be set to a specific stream-rate that a user sets.
\gls{fl} clients run a custom PyTorch dataloader that subscribes to a topic to collect the corresponding data.
With a single publisher process running, we plot the effective stream-rate achieved under slow to fast streams in Figure~(\ref{fig:multistreamrates}) and see how accurately we can simulate streaming data on a client.
A target stream-rate of 32 is closely met even while serving 16 concurrent clients with a single producer process on the central server, as shown in Figure~(\ref{fig:multiclientstreams}).
%\vspace{-0.7em}

\subsubsection{Privacy-Preserving features}

\begin{table}
  \centering
  \begin{subtable}[t]{0.48\columnwidth}
    \centering
    \begin{tabular}{|c|c|c|}
      \hline
      \multirow{2}{*}{\centering \bfseries \acrshort*{dnn}} & \multicolumn{2}{c|}{\bfseries \acrshort*{dp} Acc. (\%)} \\
      \cline{2-3}
            & \bfseries $\epsilon$=1 & \bfseries $\epsilon$=10 \\
      \hline
      Res.  & 97.98                  & \textbf{98.06} \\
      \hline
      VGG   & 24.1                   & \textbf{28.6} \\
      \hline
      Alex. & 16.31                  & \textbf{39.54} \\
      \hline
      Mob.  & 23.72                  & \textbf{58.8} \\
      \hline
    \end{tabular}
    \caption{\acrfull{dp}}
    \label{tab:diffprivacy}
  \end{subtable}
  \hfill
  \begin{subtable}[t]{0.48\columnwidth}
    \centering
    \begin{tabular}{|c|c|c|c|}
      \hline
      \multirow{2}{*}{\centering \bfseries \acrshort*{dnn}} & \multicolumn{3}{c|}{\bfseries Compute cost (sec.)} \\
      \cline{2-4}
            & \bfseries \acrshort*{dp} & \bfseries \acrshort*{he} & \bfseries \acrshort*{sa} \\
      \hline
      Res.  & 1.45                     & 68.72                    & 229.6 \\
      \hline
      VGG   & 14.4                     & 786                      & 2.3e3 \\
      \hline
      Alex. & 6.9                      & 458.7                    & 1.1e3 \\
      \hline
      Mob.  & 1.2                      & 29.8                     & 83.3 \\
      \hline
    \end{tabular}
    \caption{Privacy overhead}
    \label{tab:privacyCompares}
  \end{subtable}
  \caption{Convergence quality and compute overhead associated with privacy-preserving methods \gls{dp}, \gls{he} and \gls{sa}.}
  \label{table:privacyAlgos}
\end{table}

%\textcolor{purple}{diff. privacy convergence quality with different epsilon values}
Similar to adding different communicators and compressors, users can also specify which privacy-preserving technique to use and its pertinent parameters.
For instance, we can apply \texttt{DifferentialPrivacy} within the module \texttt{src.omnifed.privacy} and specify privacy budget with $\epsilon$ and $\delta$ from the YAML configuration itself.
With this, we measure accuracy of various \gls{dnn}s as we set $\epsilon$ to [1.0, 10.0] and $\delta$=1e-5  in Table (\ref{tab:diffprivacy}).
Large $\epsilon$ implies higher privacy budget, adding lesser noise to model updates for the same iterations, achieving more accuracy than $\epsilon$=1.
For a given model and training configuration, users can also select and compare different techniques based on privacy requirements and compute budget.
We tabulate the same in Table (\ref{tab:privacyCompares}) and observe that cryptography-based privacy techniques like \gls{he} and \gls{sa} carry significant computational overhead compared to \gls{dp} (especially for larger models like VGG11 and AlexNet).
Although this outlines the need for further optimizations to accelerate \gls{he} and \gls{sa} in \gls{dnn} training, \gls{fl} developers can experiment with various privacy mechanisms based on their needs and use cases.
%\vspace{-0.7em}

\subsubsection{Custom topology with mixed-communication protocols}

\begin{figure}
  \centering
  \begin{subfigure}[t]{0.48\columnwidth}
    \centering
    \includegraphics[width=\linewidth]{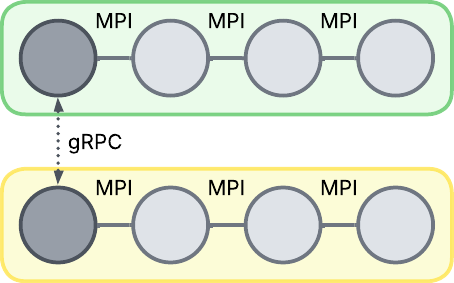}
    \caption{Topology structure}
    \label{fig:crossfacilitytopo}
  \end{subfigure}
  \hfill
  \begin{subfigure}[t]{0.5\columnwidth}
    \centering
    \includegraphics[width=\linewidth]{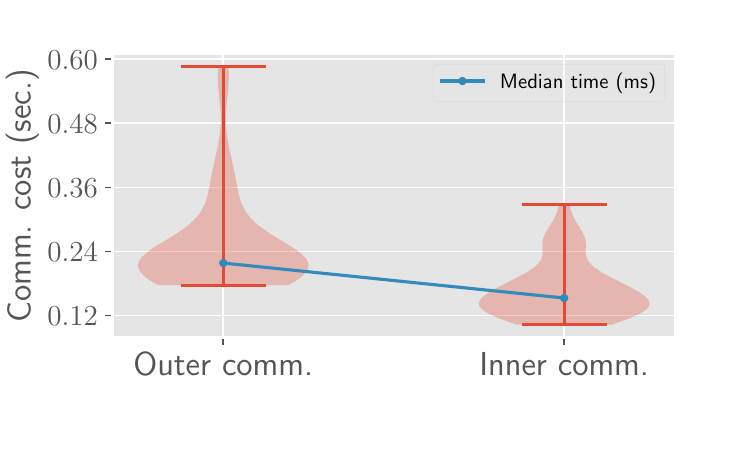}
    \caption{Outer vs.
      Inner comm.}
    \label{fig:crossfacilityCommCost}
  \end{subfigure}
  \caption{An example of cross-facility \gls{fl} with multiple communication protocols and their respective overheads.}
  \label{fig:crossfacilitytrain}
\end{figure}

With \gls{omnifed}'s modular and configurable design, users can perform cross-facility training where multiple sites (possibly geo-distributed) collaborate to learn a globally shared model.
Figure~(\ref{fig:crossfacilitytopo}) simulates the aforementioned hub-and-spoke topology where nodes within a site are densely connected with high-bandwidth links, and sparsely connected across sites over high-latency, low-bandwidth network.
Aggregation within a site can thus leverage bandwidth-optimal \gls{mpi} collectives like ring-allreduce~\cite{horovodringAR} over high bandwidths (corresponding to an inner communicator).
At the same time, cross-site communication may use \gls{grpc} over relatively slower networks (i.e., cross-facility outer communicator).
Here, inner comm. thus refers to data exchange within a sub-cluster, while outer comm. implies across sub-clusters.
With \gls{omnifed}'s modular approach, users can aggregate entire model updates over the fast, inner aggregator, and use compression (from \S \ref{subsub:compressomnifed}) \emph{only} over the slow, outer communicator (dashed line in Figure~(\ref{fig:crossfacilitytopo})).
%and/or compression techniques described in \S \ref{subsub:compressomnifed}.
%With \gls{mpi} and \gls{grpc} as local and global communicators, Figure~(\ref{fig:crossfacilityCommCost}) shows the overhead with the two protocols.
Figure~(\ref{fig:crossfacilityCommCost}) shows the overhead with \gls{mpi} and \gls{grpc} as inner and outer communicators respectively.
Inner comm. over \gls{mpi} using bandwidth-optimal allreduce is faster than global \gls{grpc} based on client-server communication.

\section{Conclusion and Future Work}\label{sec:conclusion}

In this work, we introduced \gls{omnifed}, a modular and flexible framework for \gls{fl}/\gls{cl} on edge and \gls{hpc} systems.
It enables rapid algorithm prototyping through clear separation of concerns: configuration-driven topology definition, pluggable communication protocols, and supports override-what-you-need paradigm.
It supports 10+ \gls{fl} algorithms out-of-the-box, mixed-protocol communication (combining \gls{mpi} and \gls{grpc} within a single deployment for enabling cross-facility training), pluggable privacy and compression methods.
%(to reduce communication cost).
%\gls{omnifed} supports graph-based topology definitions and streaming data simulations as well.
%Such a design allows switching between algorithms like FedAvg and FedProx with a single configuration line change, enabling systematic algorithm comparison and deployment flexibility.
%We use various models and datasets to emulate data-streaming scenarios, evaluate \gls{fl} algorithms and gradient compression techniques to compare their performance.
We emulate data-streaming scenarios, evaluate \gls{fl} algorithms and gradient compression techniques to compare their performance.
\gls{omnifed} addresses practical research needs for systematic and rapid \gls{fl} prototyping, enabling researchers to focus on design and innovation rather than infrastructure and setup complexities.
Although the current evaluation setup is limited to controlled single-facility settings, we are actively working on large-scale \gls{fl} deployment via Docker and Slurm integration.
%We're working on adding features like peer-to-peer communication, \gls{mqtt} protocol validation, and asynchronous training capabilities.
We're plan to add peer-to-peer communication, \gls{mqtt} protocol validation, heterogeneity-aware computing capabilities and integrate with optimized \gls{ml} backends like DeepSpeed~\cite{deepspeed} and TorchTitan~\cite{torchtitan} for large models, and ExecuTorch~\cite{executorch} for edge devices.
%We also plan to add heterogeneity-aware computing capabilities and integrate with optimized \gls{ml} backends like DeepSpeed~\cite{deepspeed} and TorchTitan~\cite{torchtitan} for large models, and ExecuTorch~\cite{executorch} for edge devices.
We believe such efforts will position \gls{omnifed} as a production-ready platform for collaborative research spanning over the full compute continuum.

\begin{acks}
This research used resources of the Oak Ridge Leadership Computing Facility at the Oak Ridge National Laboratory, which is supported by the Office of Science of the U.S. Department of Energy under Contract No. DE-AC05-00OR22725.
\end{acks}

\bibliographystyle{ACM-Reference-Format}
\bibliography{references}
\end{document}